\documentclass[letterpaper, 10 pt, conference]{ieeeconf}  

\IEEEoverridecommandlockouts                              

\usepackage{graphics} 
\usepackage{epsfig} 
\usepackage{times} 
\usepackage{amsmath} 
\usepackage{amssymb}  

\usepackage{graphicx}
\usepackage[utf8]{inputenc} 
\usepackage[T1]{fontenc}    
\usepackage{hyperref}       
\usepackage{url}            
\usepackage{booktabs}       
\usepackage{multirow}
\usepackage{cite}
\usepackage{svg}
\usepackage{siunitx}
\usepackage{pifont}
\newcommand{\cmark}{\textcolor{green!60!black}{\ding{51}}}
\newcommand{\xmark}{\textcolor{red!70!black}{\ding{55}}}

\usepackage{booktabs}
\usepackage[table]{xcolor}
\usepackage{makecell}

\definecolor{Gray}{gray}{0.85}

\usepackage[font=small]{caption}

\title{\LARGE \bf
\textsc{Gen2Real}: Towards Demo-Free Dexterous Manipulation by Harnessing 
Generated Video
}

\author{Kai Ye$^{1,2\ast}$, Yuhang Wu$^{2,3\ast}$, Shuyuan Hu$^{2}$, Junliang Li$^{1,2}$, Meng Liu$^{2}$, Yongquan Chen$^{1,2\dagger}$, Rui Huang$^{1,2\dagger}$
\thanks{$^{\ast}$Contributed equally to this work.}
\thanks{$^{\dagger}$Corresponding emails: {\tt \{yqchen, ruihuang\}@cuhk.edu.cn}}%
\thanks{$^{1}$The Chinese University of Hong Kong, Shenzhen. $^{2}$ Shenzhen Institute of Artificial Intelligence and Robotics for Society. $^{3}$University of California, San Diego.
}%
}

\begin{document}

\maketitle
\thispagestyle{empty}
\pagestyle{empty}

\begin{abstract}
Dexterous manipulation remains a challenging robotics problem, largely due to the difficulty of collecting extensive human demonstrations for learning. In this paper, we introduce \textsc{Gen2Real}, which replaces costly human demos with one generated video and drives robot skill from it: 
it combines demonstration generation that leverages video generation with pose and depth estimation to yield hand–object trajectories, trajectory optimization that uses Physics-aware Interaction Optimization Model (PIOM) to impose physics consistency, and demonstration learning that retargets human motions to a robot hand and stabilizes control with an anchor-based residual Proximal Policy Optimization (PPO) policy.
Using only generated videos, the learned policy achieves a 77.3\% success rate on grasping tasks in simulation and demonstrates coherent executions on a real robot. 
We also conduct ablation studies to validate the contribution of each component and demonstrate the ability to directly specify tasks using natural language, highlighting the flexibility and robustness of \textsc{Gen2Real} in generalizing grasping skills from imagined videos to real-world execution.

\end{abstract}

\section{Introduction}

Translating human intentions into physically executable robotic behaviors has long been a driving objective in embodied AI. While substantial progress has been made with rigid grippers, dexterous hands provide the versatility needed for complex tasks in unstructured settings. However, robust and generalizable control remains difficult due to high-DoF kinematics and contact-rich dynamics.

A widely adopted paradigm for dexterous manipulation learning relies on collecting large-scale demonstrations through teleoperation \cite{realdex, DIME} or motion capture \cite{Oakink2, dexycb}. These data-driven methods have enabled considerable progress, but the collection process requires specialized hardware and intensive human labor. Meanwhile, synthetic datasets generated in simulation through rule-based procedures \cite{DexGraspNet2, shao2024bimanual} are scalable, yet they often fail to capture the richness and realism of natural human interactions, leading to severe gaps when transferring policies to the real world.

An appealing alternative is to learn directly from human demonstrations in the wild \cite{ViViDex, Human2Sim2Robot, ManipTrans}, but human time is not always available when robots need to acquire new skills, and comprehensive coverage across objects, viewpoints, and tasks is difficult to guarantee. This motivates a demo-free formulation: can we acquire the benefits of human-style interaction without collecting any human demonstrations?

We answer in the affirmative by leveraging generated human demonstrations. Prior work \cite{Dreamitate, Gen2act, TASTERob} has shown that generated videos can be used to supervise low-DoF end-effectors, but directly carrying this recipe over to dexterous hands is nontrivial. Low-DoF tasks are often solvable with object-centric reasoning, where the object is guided to follow a target trajectory, whereas dexterous manipulation depends on fine-grained hand–object interactions with multi-finger coordination and contact-rich dynamics. Accordingly, we represent demonstrations explicitly as coupled hand–object interaction trajectories.

\begin{table*}[ht]
\caption{\textbf{Comparison with related methods.} Abbreviations: Gen.\ = Generated; \cmark\ = supported; \xmark\ = not supported.}
\label{tab:comparison}
\small
\centering
\resizebox{\textwidth}{!}{
\begin{tabular}{lccccc}
\toprule
 & \makecell{Demonstration \\ Data Type}  & End Effector Type & \makecell{Manipulation \\ Representation} & \makecell{Real-World \\ Deployment} & \makecell{Interaction \\ Optimization} \\
\midrule
UniPolicies \cite{UniPolicies}            & Gen. Agent Video      & Suction Gripper     & Implicit Features   & \xmark & \xmark \\
Im2Flow2Act \cite{Im2Flow2Act}            & Gen. Light Flow  & Parallel Gripper    & Object Flow         & \cmark & \xmark \\
HOPMan \cite{HOPMan}                      & Gen. Mask Video    & Parallel Gripper    & Hand-Object Mask    & \cmark & \xmark \\
ViViDex \cite{ViViDex}                    & Real Human Video      & Dexterous Hand      & Hand-Object Pose    & \cmark & \xmark \\
Human2Sim2Robot \cite{Human2Sim2Robot}    & Real Human Video      & Dexterous Hand      & Object Pose         & \cmark & \cmark \\
Maniptrans \cite{ManipTrans}              & Real Human Video      & Dexterous Hand      & Hand-Object Pose    & \cmark & \xmark \\
Dreamitate \cite{Dreamitate}              & Gen. Human Video      & Tools               & Tool Pose           & \cmark & \xmark \\
Gen2Act \cite{Gen2act}                    & Gen. Human Video      & Parallel Gripper    & Implicit Features   & \cmark & \xmark \\
TASTE-Rob \cite{TASTERob}                 & Gen. Human Video      & Parallel Gripper    & Hand Pose           & \xmark & \xmark \\
\midrule
\textbf{\textsc{Gen2Real} (Ours)}                  & Gen. Human Video      & Dexterous Hand      & Hand-Object Pose    & \cmark & \cmark \\
\bottomrule
\end{tabular}}
\end{table*}

Nevertheless, directly using generated human demonstrations is far from sufficient. Generated videos frequently contain artifacts, unstable trajectories, and physically implausible interactions. Furthermore, parsing accurate hand and object trajectories from raw video remains imperfect, compounding the problem. To address these challenges, we introduce a Physics-aware Interaction Optimization Model (PIOM), which optimizes parsed trajectories by enforcing geometric consistency, contact plausibility, and temporal smoothness, producing physically optimized trajectories.

Even with physically consistent motion, human hand trajectories cannot be directly replayed by robotic embodiments due to structural differences \cite{ViViDex, ManipTrans}. Thus, we employ constrained kinematic retargeting to map the optimized human trajectory into feasible robot action. To further close the gap between visually plausible motion and dynamically executable action, we adopt reinforcement learning techniques, specifically an anchor-based residual Proximal Policy Optimization (PPO) policy, which ensures stable performance in real-world deployment.

We formalize the task of demo-free dexterous manipulation: learning physically executable, robot-ready policies without any human-collected demonstrations. In our setting, a generated human video is the sole supervision signal, which reduces data-collection cost but also introduces challenges in realism, perception noise, and embodiment mismatch. To address these challenges, our contributions are threefold:
\begin{itemize}
    \item We present \textbf{\textsc{Gen2Real}}, a demonstration-free framework that translates natural language human intent into executable dexterous-hand policies.
    \item We introduce \textbf{PIOM}, which optimizes hand–object pose trajectory parsed from generated videos into physically optimized pose trajectory.
    \item We empirically validate the framework in both simulation and real-world experiments, demonstrating its feasibility.
\end{itemize}

\section{Related Work}

\subsection{Dexterous Manipulation}

In the field of robotic manipulation, the operation methods based on conventional rigid grippers have become relatively mature \cite{GraspAnything,huang2024rekep}.
However, with the increasing complexity and diversity of manipulation tasks in real-world robotic applications, dexterous hands, which are capable of performing fine and complex manipulation tasks \cite{zhang2024catch,GraspMulObj}, have attracted growing attention.
Traditional dexterous manipulation methods are mostly based on analytical or heuristic approaches \cite{li2024geometric,GraspWYWant}.
These approaches typically perform well only in specific scenarios and lack the generalization capability required for diverse real-world environments.
Data-driven methods—reinforcement learning \cite{DextrAHG,crossEmbodimentRL} and deep learning \cite{2025DexgraspAnything,dro,DexGraspNet2} for grasp and policy synthesis—broaden applicability. However, scalable data acquisition remains the bottleneck: real-world datasets typically rely on teleoperation or motion-capture infrastructure \cite{realdex,DIME, Oakink2,dexycb}, whereas large-scale simulation data are rule-driven \cite{DexGraspNet2,shao2024bimanual} and diverge from human interaction patterns, exacerbating sim-to-real gaps.

\subsection{Video Imitation Learning}

Imitation learning enables robots to accomplish tasks without complex planning or reward engineering by learning from agent \cite{PerceiverActor,ALOHA} or human \cite{ViViDex,Human2Sim2Robot,ManipTrans,Ditto,OKAMI, LearningContinousGrasp} video demonstrations.
Two main paradigms have emerged.
Direct visual imitation adopts end-to-end policy learning, mapping visual input from videos to robot actions \cite{MimicPlay,zhu2025learning}, which reduces manual design but demands large, diverse datasets to cope with high-dimensional visual features.
Structured-representation methods first extract keypoints, trajectories, masks or light flows from videos \cite{ViViDex, HOPMan, Im2Flow2Act}, and then learn policies based on these representations.
To reduce data demands, few/one-shot strategies leverage curricula and domain randomization \cite{shi2025learning,Dextransfer}.
However, these pipelines still hinge on the availability and coverage of human videos, which motivates a demo-free alternative using generated demonstrations.

\begin{figure*}[ht]
    \centering
    \includegraphics[width=1\linewidth]{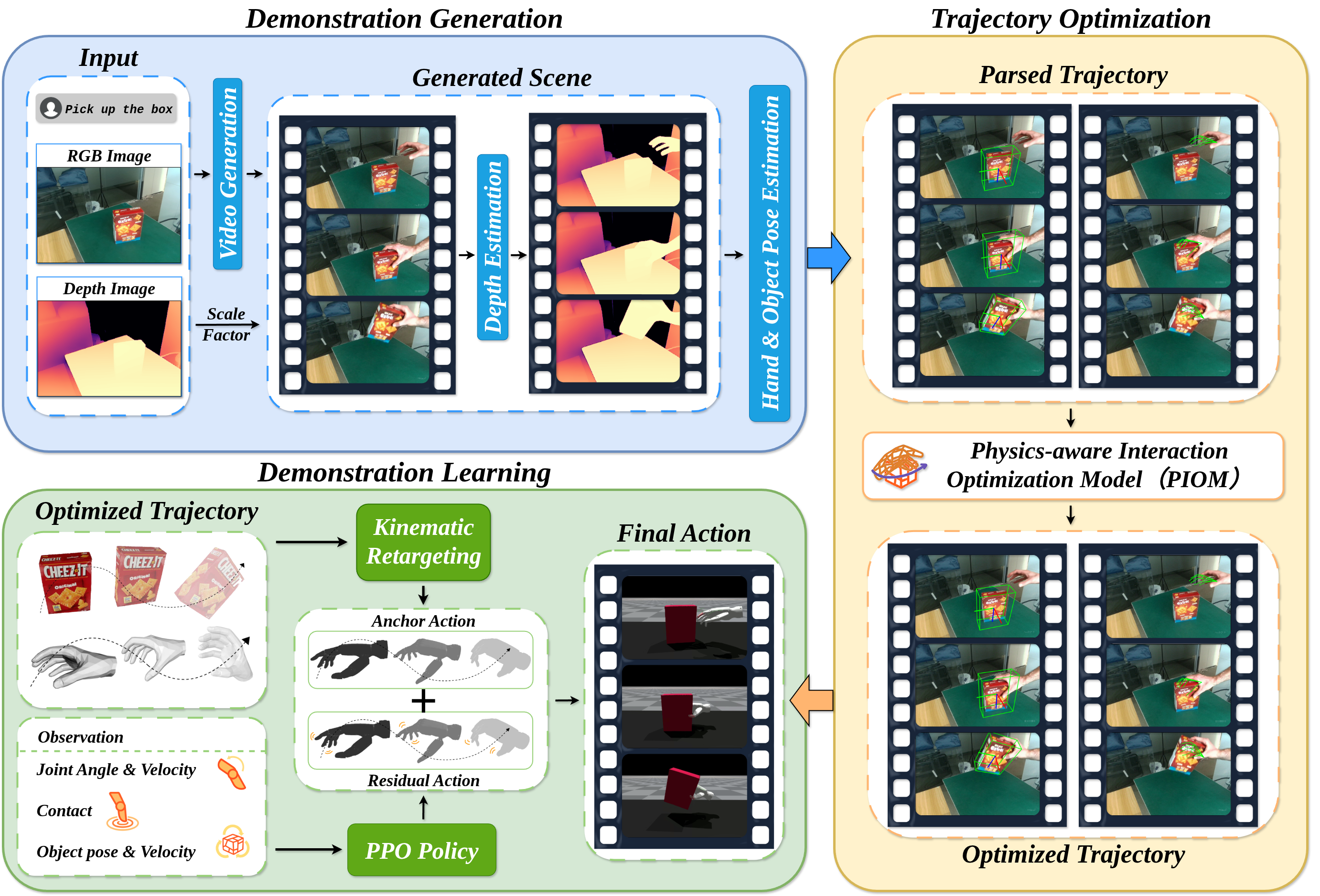}
    \caption{\textbf{\textsc{Gen2Real} pipeline.} From a single image and text, we generate video, estimate depth and poses, optimize with PIOM, retarget to the robot, and apply a residual PPO policy with anchor action, achieving generation to real transfer without human demos.}
    \label{fig:whole_pipeline}
\end{figure*}

\subsection{Generated Demonstrations for Manipulation}
Recent advances in text-to-video generation \cite{Dynamicrafter,sora} have introduced a novel paradigm for producing robotic demonstration data \cite{Position,UniSim}.
Emerging pipelines from synthetic video to policy learning fall into two families. (i) Robot-centric generation: videos of the robot itself are synthesized and used as targets for imitation learning, training policies that map visual inputs to actions to match frame-by-frame goals \cite{VLP,UniPolicies,HiP,SuSIE}. 
(ii) Human-centric generation: videos of humans are synthesized and then translated into robot supervision either by enforcing object-centric motion constraints derived from optical flow, keypoints, or masks \cite{Gen2act,HOPMan,Im2Flow2Act}, or by parsing and aligning end-effector poses with the generated motions \cite{Dreamitate}. 
These two types of approaches, built upon low-DoF end-effectors, have demonstrated the feasibility and potential of applying generative video models to robotic manipulation. 
In the more complex state space of dexterous manipulation, some recent studies have started to explore how to improve the physical consistency of hand-object interactions in generated videos \cite{TASTERob,ManiVideo}, yet no existing work has successfully achieved dexterous manipulation through generated demonstration videos (see Table~\ref{tab:comparison}).

\section{Method}

In this section, we present a novel framework for learning dexterous manipulation without human demonstrations, which we term \textsc{Gen2Real}. Our core insight is to leverage the power of large vision language models to generate diverse human manipulation demonstrations, which can then be parsed, optimized, and adapted to make them suitable for robotic execution. Fig. \ref{fig:whole_pipeline} provides an overview of our proposed method.

\subsection{Problem Definition}
\label{problem}

In this framework, we assume that the environment contains a target rigid object $\mathcal{O}$ and a robot agent $\mathcal{R}$.
At the beginning of each task, the robot receives a visual observation $I_0 \in \mathbb{R}^{3\times H\times W}$, a depth image $D_0 \in \mathbb{R}^{H\times W}$, and a natural language instruction $\mathbb{T}$. It is then required  to plan an executable action sequence $\tilde\tau^{\mathcal{R}} = \{\tilde a_1, \tilde a_2, \dots, \tilde a_T\}$ such that the robot completes the task within a finite number of steps.

\subsection{Demonstration Generation}
\label{demo_gen}

A single RGB environment image $i_0$ together with a natural-language task instruction $\mathbb{T}$ is first fed into the image-to-video generator Kling AI
\footnote{https://app.klingai.com/global/} to synthesize a task-relevant hand–object interaction video $v_I =\{I_t\}_{t=1}^T \in \mathbb{R}^{T \times H\times W\times 3}$.
The resulting frames are passed to DAV \cite{dav} to predict a dense depth sequence $v_D = \{D_t\}_{t=1}^{T} \in \mathbb{R}^{T \times H \times W}$.
Because the generated depth video is scale-invariant, we perform a global depth-scale alignment across the entire sequence.
To extract coarse hand trajectories, the hand parsing module HaMeR \cite{HaMeR} processes the RGB frames $I_t$ and outputs for each frame a wrist pose $\bar w_{t} \in \mathrm{SE}(3)$ and the hand joint angle $\bar j_{t}$ defined by MANO \cite{mano}, jointly denoted $\bar \tau^{\mathcal{H}} = \{(\bar w_{t}, \bar j_{t})\}^{T}_{t=1}$.
Since HaMeR takes RGB image as input, we additionally leverage the predicted depth sequence $v_D$ to construct per-frame hand point clouds and perform Iterative Closest Point registration to align them, thereby refining the hand poses.
In parallel, the object parsing module FoundationPose \cite{foundationpose} fuses the RGB video $v_I$ with depth video $v_D$ to predict for each frame an object pose $\bar p_{t} \in \mathrm{SE}(3)$, giving $\bar \tau^{\mathcal{O}} = \{\bar p_{t}\}^T_{t=1}$.

\begin{figure*}[t]
    \centering
    \includegraphics[width=1\linewidth]{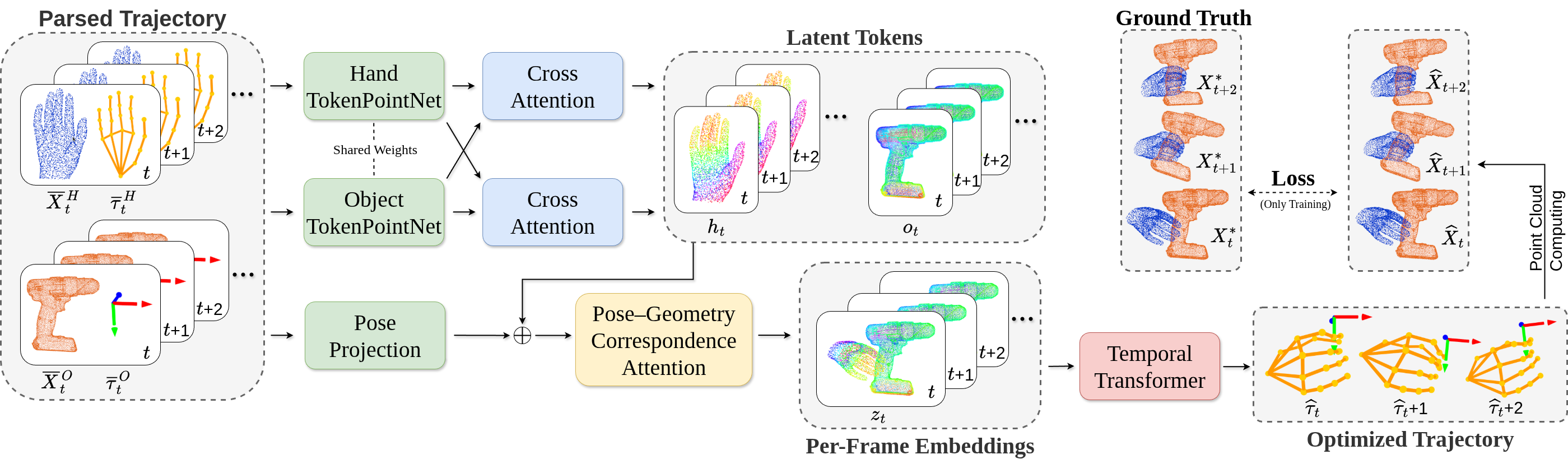}
    \caption{\textbf{Architecture of the PIOM.} The model refines noisy hand–object trajectories with attention and temporal transformers, producing optimized and physically consistent trajectories for dexterous manipulation.}
    \label{figures:PIOM}
\end{figure*}

\subsection{Physics-aware Interaction Optimization Model}
\label{method_PIOM}

Unlike trajectories recovered from sensor-verified real-video datasets, directly using $\bar \tau^{\mathcal{H}}$ and $\bar \tau^{\mathcal{O}}$ parsed from generated content often leads to high-frequency jitter, spatiotemporal misalignment, and physically implausible motion states.
To overcome the issues mentioned above, we propose PIOM, which optimizes the trajectory into physically consistent and executable sequences.

\subsubsection{Architecture}
Given a parsed interaction trajectory $\bar \tau ={(\bar w_t, \bar j_t,\bar p_t)}_{t=1}^T$.
PIOM, whose overall architecture is shown in Fig.~\ref{figures:PIOM}, first encodes the hand $\bar{X}^\mathcal{H}$ and object $\bar{X}^\mathcal{O}$ point clouds with two TokenPointNet backbones that share weights, producing latent tokens
$\mathbf{h}_t\in\mathbb{R}^{d}$ and $\mathbf{o}_t\in\mathbb{R}^{d}$, where $d$ denotes the hidden feature dimension.
The TokenPointNet employs Farthest Point Sampling to select $K$ representative points, followed by a ball query to gather local neighborhood features, where $K$ is a predefined number of keypoints.

A cross–attention block lets $\mathbf{h}_t$ attend to $\mathbf{o}_t$ (and vice-versa), injecting grasp-specific correspondences such as candidate contact pairs.  
The pose features  
$\bar \tau_t$  
are linearly projected into the same space and concatenated with the point-cloud tokens to form
$\mathcal{Z}_t=\{\mathbf{h}_t,\mathbf{o}_t,\bar \tau_t\}$.  
We then design Pose–Geometry Correspondence Attention (PGCA), which performs intra-frame self-attention over $\mathcal{Z}_t$ and fuses local geometry with instantaneous kinematics, yielding per-frame embeddings
$\mathbf{z}_t\in\mathbb{R}^{(K^\mathcal{H}+K^\mathcal{O}+1) \times d}$. 

All time steps are then processed by a temporal transformer $\mathcal{T}_{\theta}$ with positional encoding and stacked transformer encoder layers, producing the final outputs for the hand and object: $(\hat{\tau}^\mathcal{H},\hat{\tau}^\mathcal{O})=
\mathcal{T}_{\theta}\bigl(\mathbf{z}_1,\dots,\mathbf{z}_T\bigr).$

\begin{figure}[t]
    \centering
    \includegraphics[width=1\linewidth]{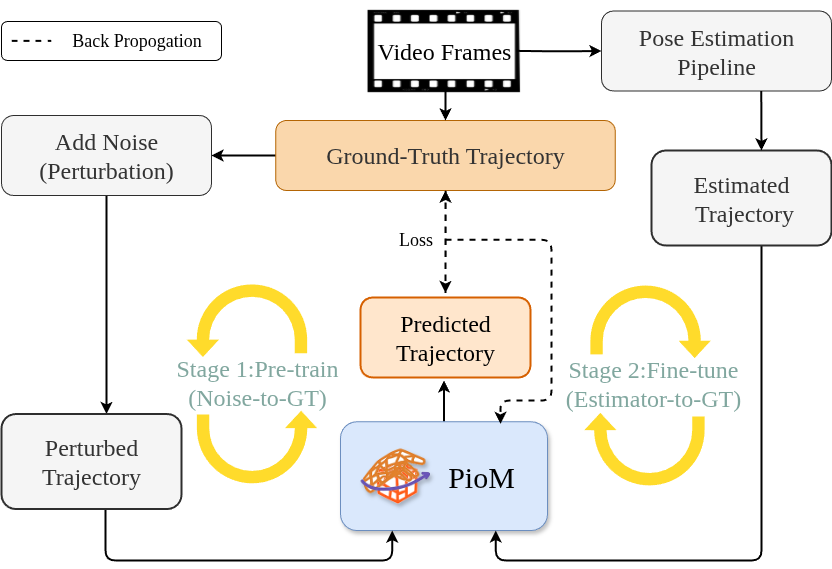}
    \caption{\textbf{Training strategy of the PIOM.} A two-stage regime is adopted: Stage 1 pre-trains the model by recovering ground-truth trajectories from perturbed inputs, and Stage 2 fine-tunes it to correct errors from the pose-estimation pipeline.}
    \label{figures:PIOM training}
\end{figure}

\subsubsection{Training strategy}
\label{train_strategy}
We adopt a two-stage supervised training regime comprising pre-training and fine-tuning (A detailed overview of our training strategy is shown in Fig. \ref{figures:PIOM training}).
In the pre-train stage, we obtain the ground-truth hand-object pose trajectory, denoted as $\tau^\ast$. 
To synthesize noisy inputs, we apply several perturbation rules that mimic the common error patterns introduced by demonstration generation, including Gaussian noise on poses, accumulated drift, relative pose bias with slight jitter, and noise on the hand joint angle.
In the fine-tune stage, the videos that correspond to the ground-truth data are processed by the pose-estimation pipeline described in Sec.\ref{demo_gen}, producing raw estimates  
$\bar \tau$.
These are fed as inputs while the original GT  serves as supervision:  
$\mathcal L\bigl($PIOM$(\bar \tau),\tau^\ast)$, thereby explicitly adapting the model to real-world estimation errors.

\subsubsection{Loss Function}
Let the batch size be $B$, the time length be $T$; the object and hand point clouds contain $M$ and $V$ points, respectively.
For each batch index $b \in [1, B]$ and timestep $t \in [1, T]$, define the valid frame mask $m_{bt} \in \{0, 1\}$. 
During the loss computation stage, we compute the hand and object point clouds $x_{bt}$ from the wrist pose $w_{bt}$, hand joint angles $j_{bt}$, and object pose $p_{bt}$ together with the corresponding meshes $\mathcal{M}^\mathcal{H}$ and $\mathcal{M}^\mathcal{O}$. We further denote the predicted and ground-truth point sets as:
\[
\begin{aligned}
\widehat{X}^{\mathcal{O}}_{bt} &= \bigl\{\widehat{x}^{\mathcal{O}}_{bti}\bigr\}_{i=1}^{M}, &
X^{\mathcal{O}\ast}_{bt} &= \bigl\{x^{\mathcal{O}\ast}_{bti}\bigr\}_{i=1}^{M}, \\
\widehat{X}^{\mathcal{H}}_{bt} &= \bigl\{\widehat{x}^{\mathcal{H}}_{bti}\bigr\}_{i=1}^{V}, &
X^{\mathcal{H}\ast}_{bt} &= \bigl\{x^{\mathcal{H}\ast}_{bti}\bigr\}_{i=1}^{V}.
\end{aligned}
\]

\paragraph{Reconstruction Loss}  The reconstruction loss measures the overall geometric discrepancy between the predicted trajectory and the ground-truth trajectory. It consists of a point cloud reconstruction term for both the object and the hand, and a joint angle term for the hand.
The object point cloud reconstruction term $\mathcal{L}^{\mathcal{O}}_{\mathrm{PC}}$ is defined as ($\mathcal{L}^{\mathcal{H}}_{\mathrm{PC}}$ is computed in the same way):
\begin{equation}
    \mathcal{L}^{\mathcal{O}}_{\mathrm{PC}} = \frac{1}{\sum_{b,t} m_{bt} M} \sum_{b,t} m_{bt} \sum_{i=1}^{M} \left\| \hat{X}^{\mathcal{O}}_{bti} - X^{\mathcal{O}\ast}_{bti} \right\|_2
\end{equation}

The hand joint angle reconstruction term $\mathcal{L}_{\mathrm{JA}}$ is defined as:
\begin{equation}
\mathcal{L}_{\mathrm{JA}} = \frac{1}{\sum_{b,t} m_{bt}} \sum_{b,t} m_{bt} \left\| \hat{j}_{bt} - j_{bt}^{*} \right\|_2^2 
\end{equation}

The overall reconstruction loss is then a weighted sum of these components:
\begin{equation}
\mathcal{L}_{\mathrm{rec}}
= \lambda^{\mathcal{O}} \mathcal{L}^{\mathcal{O}}_{\mathrm{PC}}
+ \lambda^{\mathcal{H}} \mathcal{L}^{\mathcal{H}}_{\mathrm{PC}}
+ \lambda_{\mathrm{JA}} \mathcal{L}_{\mathrm{JA}}
\end{equation}

\begin{figure*}[ht]
    \centering
    \includegraphics[width=1\linewidth]{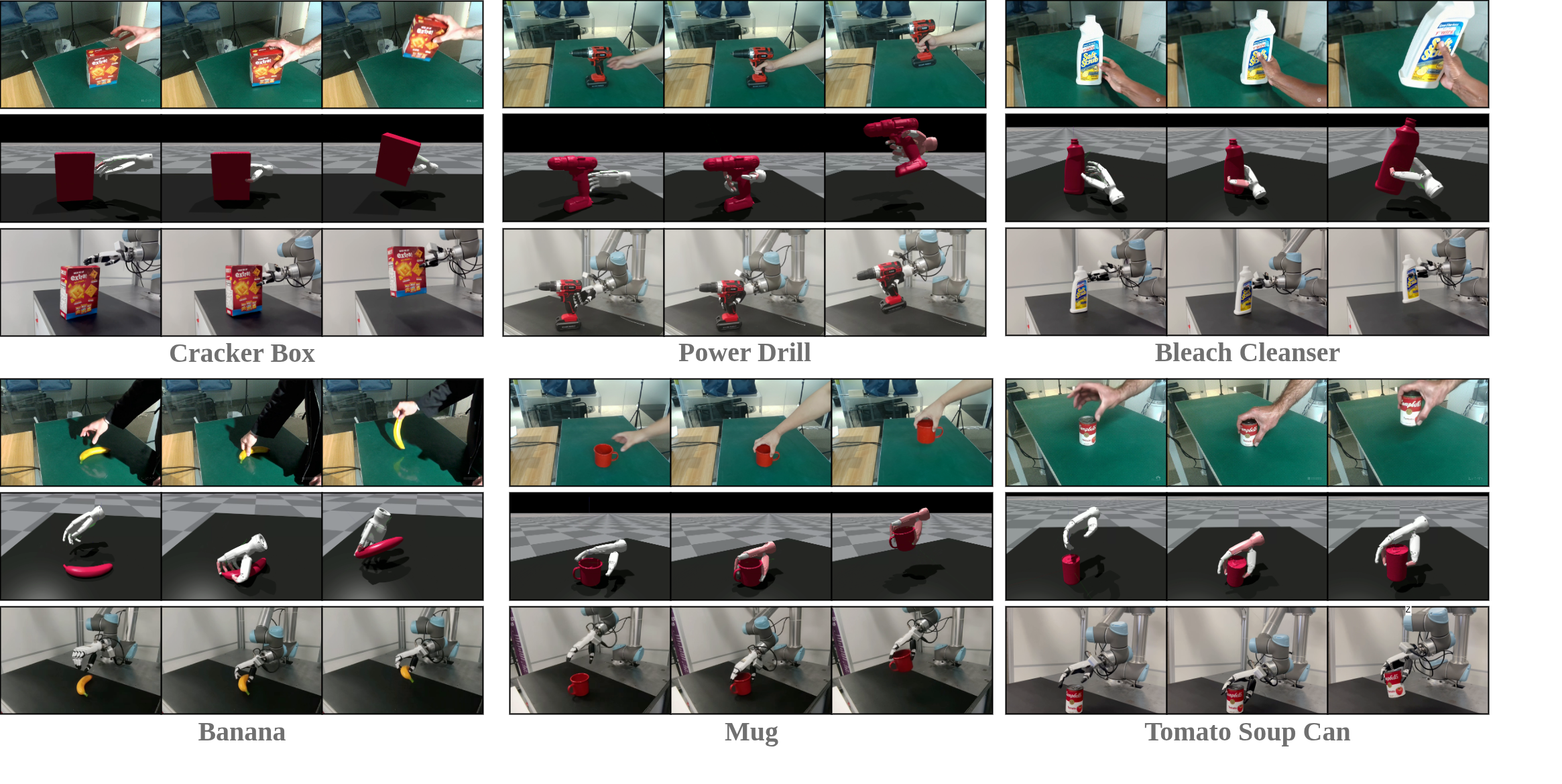}
    \caption{\textbf{Qualitative cross-domain grasping results on six YCB objects.} For each object, columns show (top) generated human hand–object frames, (middle) execution in simulation of an RL policy trained from PIOM optimized trajectories, and (bottom) execution on a real robot, demonstrating transfer through the \textsc{Gen2Real} pipeline to both sim and real.}
    \label{fig:per_object_qual}
\end{figure*}

\paragraph{Smoothness Loss} 

The smoothness loss encourages temporally consistent motion by penalizing abrupt changes in rotation and velocity between consecutive frames.
For rotation, the similarity between adjacent frames is quantified by the dot product of their relative axis-angle vectors, denoted as $\alpha$. For translation, the consistency of velocity direction is measured by the cosine similarity between consecutive velocity vectors, denoted as $\gamma$. The overall smoothness loss integrates penalties for both rotational and translational inconsistency across the sequence:
\begin{equation}
\mathcal{L}_{\mathrm{smooth}} = \sum_{b,t} m_{bt} \left( \max(0, -\alpha) + \max(0, -\gamma) \right).
\end{equation}

\paragraph{Penetration Loss}

The penetration loss penalizes the intersection between the hand and object point clouds by measuring the signed distances from each hand vertex to the nearest point on the object surface. Let \( \delta_{bti} \) denote the signed distance from a hand point \( \hat{x}^{\mathcal{H}}_{bti} \) to the closest object point \( \hat{x}^{\mathcal{O}}_{bti} \).

\begin{equation}
  \mathcal{L}_{\mathrm{pene}} = \frac{1}{\sum_{b,t} m_{bt} V} \sum_{b,t} m_{bt} \sum_{v=1}^{V} max(0, -\delta_{bti})  
\end{equation}

Overall, we define the final training objective as the weighted sum of the three losses:
\begin{equation}
\mathcal{L}_{\mathrm{total}}
= \lambda_{\mathrm{rec}}\,\mathcal{L}_{\mathrm{rec}}
+ \lambda_{\mathrm{smooth}}\,\mathcal{L}_{\mathrm{smooth}}
+ \lambda_{\mathrm{pene}}\,\mathcal{L}_{\mathrm{pene}} .
\end{equation}

\subsection{Demonstration Learning}
\label{demo_learn}

Although PIOM optimizes the parsed hand-object trajectory to ensure spatial-temporal consistency, structural differences between human and robotic hands render direct trajectory execution infeasible.
To address this challenge, we introduce a two-step demonstration learning process comprising kinematic retargeting and anchor-based residual policy learning.

\subsubsection{Kinematic Retargeting}

We obtain the retargeted robot trajectory by mapping the refined human demonstration \(\hat{\tau}^{\mathcal{H}}\) to the robot’s joint and wrist poses: $\hat{\tau}^{\mathcal{R}} = \{(\hat{w}_{t}, \hat{j}_{t})\}_{t=1}^{T} \in \arg \min J(w,j ;\hat{\tau}^{\mathcal{H}})$.
where the objective combines fingertip alignment, wrist orientation, and joint-angle similarity:
\begin{equation}
J(w,j ;\hat{\tau}^{\mathcal{H}}) = \bigl\lVert q - \hat{q}\bigr\rVert^2
      + \lambda_o\bigl\lVert log(R(w)R(\hat{w})^T)\bigr\rVert^2
      + \lambda_a\bigl\lVert j-\hat{j}\bigr\rVert^2
\end{equation}
where $q$ represents fingertip position, and $R(\cdot)$ denotes the rotation matrix extracted from a wrist pose. 

\begin{table*}[t]
  \centering
  \caption{\textbf{Per-object grasping success rates in simulation across six YCB objects.}}
  \label{tab:merged_sr}
  \small
  \setlength{\tabcolsep}{6pt}
  \renewcommand{\arraystretch}{1.15}
  \begin{tabular}{lccccccc}
    \toprule
    \textbf{Object} & Tomato Soup Can & Power Drill & Bleach Cleanser & Cracker Box & Banana & Mug & \textbf{Mean} \\
    \midrule
    \textbf{Sim SR (\%)} & 72.5 & 70.7 & 88.3 & 86.8 & 69.2 & 76.0 & \textbf{77.3} \\
    \bottomrule
  \end{tabular}
\end{table*}

\subsubsection{Anchor-Based Residual Policy}

\begin{table}[hb]
  \centering
  \caption{\textbf{Component ablation of \textsc{Gen2Real}.}}
  \label{tab:ablation_fixed_modules}
  \small
  \setlength{\tabcolsep}{4pt}
  {\rowcolors{2}{gray!20}{white}%
  \resizebox{\linewidth}{!}{%
    \begin{tabular}{l c c c c}
      \toprule
      \hiderowcolors
      & \multicolumn{3}{c}{\textbf{Components}} & \\
      \cmidrule(lr){2-4}
      \textbf{Model Variant } & \textbf{PIOM} & \textbf{Residual} & \textbf{Retargeting} & \textbf{SR} \\
      \midrule
      \showrowcolors
      No PIOM                 & \xmark & \cmark & \cmark & 0\% \\
      No PIOM (+Offset) & \xmark & \cmark & \cmark & 2.1\% \\
      No Residual             & \cmark & \xmark & \cmark & 7.5\% \\
      No Retargeting          & \cmark & \cmark & \xmark & 36.1\% \\
      \addlinespace[2pt]
      \textbf{\textsc{Gen2Real}} & \cmark & \cmark & \cmark & \textbf{77.3\%} \\
      \bottomrule
    \end{tabular}%
  }}
\end{table}

Although kinematic retargeting has provided a mapped trajectory, it is still unstable in real task deployment due to a lack of kinematic interaction knowledge. Therefore, we train a policy $\pi_{\theta}(a_t|s_t)$ via PPO to apply physics-aware residual corrections to anchor $\hat\tau^{\mathcal{R}}$.  The action in timestep $t$ is defined as $a_t = (j_t, w_t)$ (details in Sec.~\ref{problem}). The state space is defined as:
$
s_t = \{j_t, \dot{j}_t, w_t, \dot{w}_t, p_t, \dot{p}_t, C_t, \hat\tau_t^{\mathcal{R}}, \hat\tau_t^{\mathcal{O}}\}
$.
where $\dot{j}_t \in \mathbb{R}^{\mathcal{R}_{dof}}$ is joint velocity, $\dot{w}_t \in \mathbb{R}^6$ is the wrist velocity, and $\dot{p}_t \in \mathbb{R}^6$ is object velocity. $C_t \in \mathbb{R}^{F}$ represents contact force information, where $F$ denotes the number of fingertips on hands. 

We design a general reward function that integrates multiple objectives:
\begin{equation}
\begin{aligned}
r_t ={} & \; \lambda_{object} r_t^{\mathcal{O}} 
       + \lambda_{wrist} r_t^{w} 
       + \lambda_{finger} r_t^{q} 
       + \lambda_{contact} r_t^{C}.
\end{aligned}
\end{equation}

The components are defined as follows:  
1) Object following reward $r_t^{\mathcal{O}}$: Encourages accurate object tracking by minimizing positional and velocity errors between the simulated object and its reference trajectory, i.e., $p_t \ominus \hat{p}_t$ and $\dot{p}_t - \hat{\dot{p_t}}$. Here, $\ominus$ denotes the pose difference operator, which computes the relative error between two poses in $\mathrm{SE}(3)$.
2) Wrist tracking reward $r_t^{w}$: Reduces discrepancies between the simulated and reference wrist states, including $w_t \ominus \hat{w}_t$ and $\dot{w}_t - \hat{\dot{w_t}}$.  
3) Finger joint reward $r_t^{q}$: Promotes accurate joint motion by minimizing deviations in positions $j_t - \hat{j}_t$ and velocities $\dot{j}_t - \hat{\dot{j_t}}$.  
4) Contact reward $r_t^{C}$: Encourages stable multi-finger interaction by rewarding the number of fingertips in contact with the object.

\begin{table*}[ht]
  \centering
  \caption{\textbf{PIOM evaluation and ablations.} All metrics are errors w.r.t. ground-truth trajectories (lower is better).}
  \label{tab:PIOM_ablation}
  \small
  \setlength{\tabcolsep}{4pt}
  \resizebox{\textwidth}{!}{
  \begin{tabular}{l c c c c c c c c c}
    \toprule
    & \multicolumn{3}{c}{\textbf{PIOM Modules}} & \multicolumn{6}{c}{\textbf{Errors vs. GT (}\(\downarrow\)\textbf{)}} \\
    \cmidrule(lr){2-4}\cmidrule(lr){5-10}
    \textbf{Training Variant} & \textbf{Perturb} & \textbf{Cross-Attn} & \textbf{PGCA} & \textbf{MPJPE} & \textbf{ADD-S} & \textbf{FD (hand)} & \textbf{FD (obj)} & \textbf{JK (hand)} & \textbf{JK (obj)} \\
    \midrule
    Parsed Trajectory & --     & --     & --     & 834.87 & 72.19 & 1425.02 & 221.57 & 465.90 & 33.74 \\
    \addlinespace[2pt]
    \multicolumn{10}{l}{\emph{PIOM Ablations}} \\
    \rowcolor{gray!20} No Perturbation Pretraining         & \xmark & \cmark & \cmark & 58.01  & 28.85 & 125.58  & 72.15  & 19.64  & 12.24 \\
    No Cross-Attention                  & \cmark & \xmark & \cmark & 59.78  & 29.68 & 128.11  & 79.94  & 18.09  & 10.58 \\
    \rowcolor{gray!20} No PGCA                             & \cmark & \cmark & \xmark & 61.74  & 28.65 & 140.59  & 70.67  & 17.08  & 9.18  \\
    \textbf{PIOM}                       & \cmark & \cmark & \cmark & \textbf{53.65} & \textbf{23.35} & \textbf{110.00} & \textbf{64.22} & \textbf{11.83} & \textbf{7.23} \\
    \bottomrule
  \end{tabular}}
\end{table*}

\section{Experiments}
\label{sec:experiments}

We verify our idea through the following three parts: (1) Grasping Experiments: per-object grasping conducted in both simulation and on hardware (Sec. \ref{grasp_result}), together with qualitative language-conditioned grasping experiments (Sec. \ref{lang_grasp_result}). (2) \textsc{Gen2Real} Analysis: ablation studies of individual components (Sec. \ref{sec:gen2real_ablation}). (3) PIOM Analysis: quantitative trajectory optimization results (Sec. \ref{PIOM_result}), with targeted ablations to probe design choices (Sec. \ref{PIOM_ablation}).

\subsection{Grasping Experiments}
\label{sec:grasping}

\subsubsection{Setup and Metrics}
We evaluate grasping performance in both simulation and the real world. Six YCB~\cite{Ycb} objects with distinct geometries—\emph{Tomato Soup Can}, \emph{Power Drill}, \emph{Bleach Cleanser}, \emph{Cracker Box}, \emph{Banana}, and \emph{Mug}—are selected to validate the feasibility of our approach. In simulation, we use \emph{IsaacGym}~\cite{isaacgym} as the environment. Each trial is initialized with randomized object poses, and a rollout is considered successful if it satisfies all of the following criteria: rotation error $<30^\circ$, translation error $<3cm$, mean per-joint position error $<8cm$, and mean per-fingertip position error $<6cm$. For each object, we conduct $1000$ trials to obtain statistically reliable results.
In the real world, we employ the \emph{Inspire Hand} mounted on a \emph{UR5} robotic arm. In simulation we control $12$ DoF, while in the physical setup only $6$ DoF are available, for which we apply a simple mapping.

\subsubsection{Results}
\label{grasp_result}
Experiments on six YCB objects show that \textsc{Gen2Real} achieves a mean success rate of 77.3\% in simulation (Table~\ref{tab:merged_sr}), and coherent executions on the physical platform further validate transfer from generated video to real robots (Fig.~\ref{fig:per_object_qual}). These results highlight three strengths: first, a scalable demo-free pipeline in which a single intent instruction for any new object suffices to synthesize demonstrations and automatically produce executable hand-object interaction trajectories without human demonstrations or object-specific annotations; second, cross-object consistency and reusability, since one architecture, loss design, and set of hyperparameters yield stable grasps across the six objects and repeated executions of the same task require no additional training; third, a closed loop from human intention to physically executable control, showing that high-level intent can be converted into reusable and deployable dexterous grasping behaviors without collecting any real demonstrations.

\subsubsection{Language-Conditioned Grasping}
\label{lang_grasp_result}
\begin{figure}[h]
    \centering
    \includegraphics[width=1\linewidth]{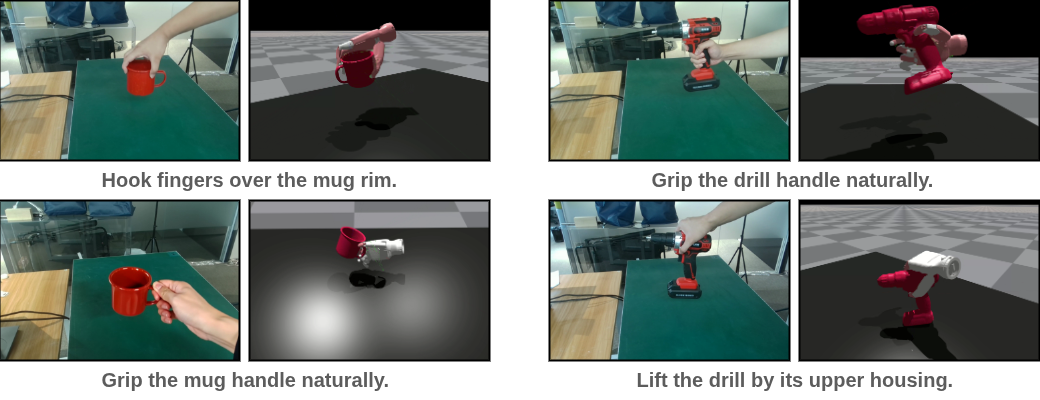}
    \caption{\textbf{Language-conditioned grasping.} Two prompts per object yield distinct grasp strategies.}
    \label{fig:language_multi_grasp}
\end{figure}

As shown in Fig.~\ref{fig:language_multi_grasp}, we issue two distinct natural-language commands for each of two objects—the \emph{mug} and the \emph{power drill}—describing markedly different grasp strategies. Conditioned on each instruction, \textsc{Gen2Real} generates the corresponding demonstration trajectories and then reproduces the specified grasps in simulation. This highlights our method's ability to accommodate diverse intents and map language to contact-rich manipulation behaviors.

\subsection{Gen2Real Analysis}

\subsubsection{Setup and Metrics}
We conduct ablation studies on three key components of \textsc{Gen2Real}. Each ablation setting was evaluated on all objects listed in Table \ref{tab:merged_sr}, and the mean performance was reported. For the removal of PIOM, we designed two variants. This is because the parsed trajectories, without optimization, typically exhibit large deviations, making it nearly impossible for the subsequent modules to succeed. To better highlight the fine-grained contribution of PIOM, we additionally introduced a variant where the parsed trajectories were corrected with simple offset adjustments. Beyond this, we also examined the effects of removing the residual strategy and the retargeting module.

\subsubsection{Ablation results}
\label{sec:gen2real_ablation}

Based on the ablation results in Table~\ref{tab:ablation_fixed_modules}, removing any of the three key modules leads to significant performance degradation, with PIOM serving as a prerequisite for the downstream components. Specifically, when PIOM is removed (while keeping retargeting and residual policy), the success rate drops from 77.3\% to 0\%, and even with a simple offset adjustment it only reaches 2.1\%, indicating that the raw parsed trajectories are too noisy for subsequent modules to function. When PIOM and retargeting are preserved but the residual policy is removed, we observe that, without the guidance of an anchor trajectory, training often stagnates at mimicking the initial actions, causing the policy to fail in maintaining stable contact and object control in the later phases of execution, yielding only a 7.5\% success rate. When PIOM and the residual policy are preserved but retargeting is removed, the success rate rises to 36.1\%; however, we find that the successful cases are mostly tasks with low requirements for precise contact points, while tasks that demand accurate fingertip placement almost always fail. Only the full configuration restores the success rate to 77.3\%, fully validating the effectiveness of all three modules.

\subsection{Physics-aware Interaction Optimization Model Analysis}
\label{sec:PIOM Analysis}

\subsubsection{Setup and metrics}
To evaluate the effectiveness of PIOM as a trajectory optimizer, we conduct experiments on the DexYCB \cite{dexycb} dataset, which provides a standard subject-level split for training and testing. All metric computations are performed on the test set. For each scene and viewpoint, we first obtain the GT from DexYCB. We feed images into the pose-estimation pipeline (see Sec.~\ref{demo_gen}) to generate a parsed hand–object interaction trajectory, which is subsequently optimized by PIOM. Each metric reported in the table is computed by comparing the optimized trajectory against the GT.
Our evaluation metrics cover both geometric accuracy and temporal consistency. Specifically, the Mean Per Joint Position Error (MPJPE)~\cite{ionescu2013human3} measures hand-joint positional error; the Average Distance of Model Points for Symmetric objects (ADD-S)~\cite{kleeberger2019large} evaluates object-pose accuracy while accounting for symmetry; and the Fréchet Distance (FD) captures trajectory-level similarity of hand and object motions in trajectory space. In addition, we report jerk (JK) for both hand and object trajectories to quantify smoothness and the presence of high-frequency oscillations. All metrics are reported in millimeters.

\subsubsection{Training setting}
To improve generalization, we apply data augmentation that includes uniformly sampled rotations from $SO(3)$ and Gaussian global translation noise ($\sigma$ =  $0.2$~m), synchronized random scaling of hand and object, and random temporal resampling that scales sequence length by [$0.7$, $1.3$] , and nearest-neighbor resampling for the validity masks. The model is optimized using Adam with a learning rate of $10^{-4}$, a batch size of $16$, and a maximum sequence length of $120$ frames. Training is conducted for $100$ epochs.

\subsubsection{Ablation}
\label{PIOM_ablation}
In the absence of perturbation pretraining (Sec. \ref{train_strategy}), all metrics deteriorate; compared with this setting, PIOM substantially reduces hand and object jerk and also improves geometry and trajectory level metrics, indicating that noise modeling that approximates real estimation errors markedly enhances temporal robustness and de-jittering while producing stable improvements across metrics. Removing cross-attention leads to the largest degradation in object-related accuracy, with noticeable declines also observed for the hand, confirming that cross-attention explicitly aligns hand and object to better capture interaction. Without PGCA, hand accuracy and trajectory similarity degrade the most, showing that intra-frame alignment of geometry and pose via self-attention is crucial for correcting fine-grained joint-level errors.

\subsubsection{Results}
\label{PIOM_result}
The results are summarized in Table~\ref{tab:PIOM_ablation}. The trajectories produced by the demonstration-generation pipeline differ significantly from the ground truth, partly because most existing models are trained on real-world data and thus degrade under generated scenes. Nevertheless, when comparing the first and last rows of the table, it becomes clear that applying PIOM yields consistent and substantial improvements across all metrics.  
For hand pose, MPJPE decreases by $93.6\%$, while for object pose, ADD-S drops by $67.7\%$, indicating more accurate alignment. On the trajectory level, hand FD is reduced by $92.3\%$ and object FD by $71.0\%$, showing that motion patterns are preserved more faithfully. Improvements in smoothness are also clear: hand JK falls by $97.5\%$, and object JK by $78.6\%$, reflecting the suppression of high-frequency artifacts and unstable contact transitions.  
Overall, PIOM not only corrects geometric errors but also enforces physically plausible and temporally coherent trajectories, thereby benefiting downstream retargeting and control.

\section{Conclusion}

We presented Gen2Real, a demo-free framework that converts a single generated human video into robot-executable dexterous-hand policies. PIOM optimizes hand–object trajectories, constrained kinematic retargeting bridges embodiment mismatch, and an anchor-based residual PPO policy closes the sim-to-real gap. In simulation, the method achieves strong success rates and transfers to coherent real-world grasps; ablations confirm each component’s contribution, and language-conditioned prompts yield diverse grasp strategies. These results demonstrate that imagined videos can be turned into physically grounded skills without any human-collected demonstrations.

\bibliographystyle{IEEEtran}
\bibliography{references}

\end{document}